\documentclass[runningheads,a4paper]{llncs}

\usepackage[T1]{fontenc}
\usepackage[utf8]{inputenc}
\usepackage[a4paper,margin=1in]{geometry}
\usepackage{graphicx}
\usepackage{booktabs}
\usepackage{array}
\usepackage{tabularx}
\usepackage{multirow}
\usepackage{amsmath}
\usepackage{amssymb}
\usepackage{url}
\usepackage{hyperref}
\usepackage{xcolor}
\usepackage{microtype}
\usepackage{enumitem}
\usepackage{tikz}
\usetikzlibrary{positioning,arrows.meta,shapes.geometric}

\hypersetup{
  colorlinks=true,
  linkcolor=blue!60!black,
  citecolor=blue!60!black,
  urlcolor=blue!60!black,
  pdftitle={Generative Ontology Induction: Domain-Agnostic Schema Discovery from Document Corpora Using Large Language Models},
  pdfauthor={Sergei Sergienko},
  pdfkeywords={ontology learning, schema induction, large language models, knowledge graphs, domain-agnostic methods, generative models}
}

\makeatletter
\renewenvironment{abstract}{%
  \list{}{\advance\topsep by0.35cm\relax\small
    \leftmargin=0pt
    \labelwidth=\z@
    \listparindent=\z@
    \itemindent\listparindent
    \rightmargin=0pt}%
  \item[\hskip\labelsep\bfseries\abstractname]}
{\endlist}
\makeatother

\begin{document}

\title{Generative Ontology Induction: Domain-Agnostic Schema Discovery
from Document Corpora Using Large Language Models}

\titlerunning{Generative Ontology Induction}

\author{Sergei Sergienko\,\href{https://orcid.org/0009-0009-5180-7115}{(ORCID: 0009-0009-5180-7115)}\thanks{An earlier version of this work was previously deposited on Zenodo (DOI: \href{https://doi.org/10.5281/zenodo.19893755}{10.5281/zenodo.19893755}).}}

\authorrunning{S. Sergienko}

\institute{Pivots Global\\
\email{ssergienko@ontology.live}}

\maketitle

\begin{abstract}
{\normalsize
Ontology engineering remains a critical bottleneck in knowledge-intensive
AI systems. Existing automated approaches either depend on predefined
schemas, operate within narrow domains, or produce unstructured outputs
unsuitable for downstream pipelines.

We introduce \textbf{Generative Ontology Induction (GOI)}, a
domain-agnostic framework that induces a \emph{generative
blueprint}-entities, dimensions, properties, relationships, and
constraints-from a corpus of examples and exports it as a typed graph
(six node types, seven edge types) in YAML/JSON. We introduce the
\textbf{Node Coverage Score}, a novel evaluation metric that measures
the fraction of structural ontology nodes (classes, properties, and
dimensions) appearing in generated outputs.

A controlled generative validation on four contrasting ontologies-a
familiar Software Services Invoice schema, a custom Job Description
Ontology, a confidential Pain-Management Clinical Visit Record
Ontology, and a Professional Services Contract \& Statement of Work
Ontology-shows that GOI-prompted generation covers 95-100\% of the
structural backbone in every case; a generic three-field template
holds at 97.8\% on the invoice schema but drops to 52.2\% on the Job
Description Ontology, 62.2\% on the Pain-Management ontology, and
78.3\% on the Professional Services Contract ontology. The
structural coverage holds regardless of how familiar the document
type is to the model.

\keywords{Ontology learning \and Schema induction \and Large language
models \and Knowledge graphs \and \mbox{Domain-agnostic} methods \and
Generative models \and Cross-functional teams}
}
\end{abstract}

\section{Introduction}
\label{sec:intro}

Modern AI systems increasingly rely on structured knowledge to ground
reasoning, guide generation, and ensure consistency across complex
workflows. Ontologies-formal representations of entities,
relationships, and constraints within a domain-serve as the backbone
for knowledge graphs, semantic search, retrieval-augmented generation
(RAG), and multi-agent orchestration. Yet ontology engineering remains a
labor-intensive, expert-driven process that creates a fundamental
bottleneck: every new domain, document type, or application requires
manual schema design, iterative refinement, and domain-specific rules.

This \textbf{structure gap} manifests across the AI landscape.
Enterprise knowledge management systems struggle to maintain consistent
schemas as document types evolve. Research teams building
domain-specific knowledge graphs invest months in ontology design before
extraction can begin. Agent frameworks that promise autonomous operation
still require hand-crafted schemas to structure their memory and
reasoning. The promise of general-purpose AI collides with the reality
that structured knowledge remains stubbornly domain-specific and
manually curated.

A particularly underexplored dimension of this gap occurs within
mixed-expertise teams: researchers, backend engineers, and frontend
developers operate with fundamentally different mental models, and lack
a shared, human-readable structural representation of the product's
knowledge layer.

Recent advances in large language models (LLMs) have begun to address
the structure gap. Zero-shot and few-shot ontology construction pipelines
demonstrate that LLMs can extract entities, induce taxonomies, and
generate schema elements without extensive training data
\cite{beliaeva2025,giglou2023llms4ol,giglou2024llms4ol_overview}. Corpus-based
methods can discover hierarchical event ontologies from open-domain text
collections~\cite{zhang2023ceo}. Interactive systems enable human-guided
schema discovery from research corpora~\cite{sadruddin2025}. Yet these
approaches share common limitations: they often require predefined
ontology seeds, operate within constrained domains, produce untyped or
inconsistently structured outputs, lack robust cross-document
canonicalization, and provide limited pathways for non-expert users to
visualize, validate, and export discovered schemas.

This paper introduces \textbf{Generative Ontology Induction (GOI)}, a
framework designed to address these gaps through four core contributions:

\textbf{First}, we formalize the concept of \emph{generative
ontology}-the structural schema that defines a document class and
enables generation of new instances-and distinguish it from
descriptive entity extraction. GOI reverse-engineers this generative
blueprint by analyzing multiple examples of the same document type,
identifying recurring dimensions, properties, relationships, and
constraints that govern the class.

\textbf{Second}, we present a complete technical architecture
comprising a universal typed graph representation (six node types, seven
edge types), a multi-document prompt that elicits generative schemas
rather than per-document entities, an automatic type-based layout
algorithm, multi-format import (OWL, RDF, YAML, JSON, Markdown), and a
dual-view YAML/JSON export whose \texttt{promptReady} markdown
rendering realizes the ``ontology drives AI pipelines'' thesis by
allowing direct injection into RAG and agent prompts without graph
traversal.

\textbf{Third}, we introduce the \textbf{Node Coverage Score}, a novel
evaluation metric that measures the fraction of structural nodes
(classes, properties, and dimensions) in an induced ontology that
appear in outputs generated from that ontology-a direct
structural-completeness signal that token-level metrics miss.

\textbf{Fourth}, we identify and address a practical
\textbf{cross-functional team communication gap}: mixed-expertise AI
product teams lack a shared, readable structural artifact, and GOI's
visual, interactive, exportable schema serves as precisely this common
artifact, positioning GOI as both a technical contribution and an
\textbf{organizational primitive}.

A controlled generative validation on four contrasting ontologies
spanning B2B billing, talent acquisition, healthcare clinical
documentation, and professional services contracting demonstrates
GOI's structurally-mandated coverage; we also discuss limitations
including LLM consistency, token-window constraints, the lack of
formal OWL axioms, and hallucination risks.

The remainder of this paper is organized as follows.
Section~\ref{sec:related} reviews related work.
Section~\ref{sec:framework} details the GOI framework.
Section~\ref{sec:metric} introduces the Node Coverage Score.
Section~\ref{sec:experiments} presents experimental results.
Section~\ref{sec:discussion} discusses implications and limitations.
Section~\ref{sec:conclusion} concludes.

\section{Related Work}
\label{sec:related}

\subsection{Zero-Shot and Few-Shot Schema Induction}

LLM-driven methods now produce cross-domain schemas and corpus-level
ontologies without heavy supervision. Corpus-based open induction uses
distant supervision and corpus-wide signals to induce hierarchical event
ontologies without direct supervision~\cite{zhang2023ceo}. Zero-shot
generation of sources synthesizes or retrieves pseudo-documents to
expose latent schema structure and then induces schema elements in a
zero-shot manner~\cite{giglou2023llms4ol}. Schema-from-text pipelines that
do not assume a predefined schema perform extraction then define and
canonicalize schema elements post-hoc, enabling domain-agnostic
operation~\cite{zhang2024edc}. Prompt-only competitive methods apply LLM
prompting-including retrieval-augmented prompts and sampling
strategies-to construct terms, types, and taxonomies across domains
without model fine-tuning~\cite{beliaeva2025}. OntoGenix~\cite{canobenito2024ontogenix}
shows that LLMs can be directly leveraged for ontology engineering from
tabular datasets, generating reusable conceptual models. Recent
benchmarking efforts such as Text2KGBench~\cite{mihindukulasooriya2023text2kgbench}
have begun to standardize evaluation of ontology-driven knowledge graph
generation from text, surfacing the need for structural-not just
token-level-evaluation criteria.

\subsection{Knowledge Graph Construction from Document Collections}

iText2KG~\cite{lairgi2024itext2kg} processes document collections
incrementally, using a graph integrator to merge entity and relation
extractions across documents into a unified knowledge graph. The system
explicitly acknowledges that unresolved semantically duplicate entities
remain a challenge. AutoClusRE~\cite{wang2024autoclusre} builds
corpus-level type tables guided by dynamically updated entity and
relation type tables, using semantic clustering to reduce duplication
across texts. Iterative zero-shot prompting
pipelines~\cite{giglou2024llms4ol_overview} offer scalable corpus-level
extraction without training examples. Graph-based event schema
induction~\cite{xu2021graph} builds event graphs and clusters them to
induce recurring event templates across documents. SAC-KG~\cite{chen2024sackg}
treats LLMs as automatic constructors of domain-specific knowledge
graphs, demonstrating that prompt-driven pipelines can scale beyond
hand-crafted extractors. At the application layer, Microsoft's
GraphRAG~\cite{edge2024graphrag} highlights how downstream
retrieval-augmented systems benefit from explicit graph-structured
knowledge over raw text indices, motivating the need for high-quality
schemas of the kind GOI induces.

\subsection{Ontology Learning with LLMs}

Lo et al.~\cite{lo2024endtoend} present an end-to-end ontology learning
pipeline with LLMs and introduce structural and semantic graph distance
measures as evaluation metrics, noting that standard token-level metrics
miss structural correctness and coherence. Beliaeva and
Rahmatullaev~\cite{beliaeva2025} survey heterogeneous LLM methods for
ontology learning, identifying canonicalization and evaluation as open
problems. Sadruddin et al.~\cite{sadruddin2025} introduce
LLMs4SchemaDiscovery, a human-in-the-loop system that converts research
questions plus a corpus into a structured schema and grounded database.
OntoKGen~\cite{abolhasani2025ontokgen} provides an adaptive iterative
Chain-of-Thought interface for ontology extraction, integrating
generated knowledge graphs into Neo4j for flexible querying and visual
exploration. NeOn-GPT~\cite{fathallah2024neongpt} combines a methodology
framework with LLM prompts to produce Turtle ontologies, finding that
LLMs reduce effort but lack procedural reasoning capabilities. OntoGPT's
SPIRES system~\cite{caufield2024spires} extracts semantic structures
according to LinkML schemas without training data, demonstrating
schema-guided extraction patterns. ReCG~\cite{yun2024recg} applies
bottom-up cluster-and-generalize with minimum description length (MDL)
to discover JSON schemas from document collections.

\subsection{Visual Ontology Editors and Accessibility}

KGraphX~\cite{hemid2024kgraphx} is an easy-to-use visual editor
for users with limited semantic web expertise, supporting drag-and-drop
graph construction, entity lookup from Wikidata and BioPortal, and
RDF/RDFS export. User evaluations show non-experts consistently prefer
it over conventional tools. WebProt\'eg\'e~\cite{tudorache2013webprotege}
provides collaborative browser-based ontology editing for distributed
teams, with change tracking and annotation threads, lowering the barrier
to participation without requiring local tool installation.
Metaphactory~\cite{haase2019metaphactory} targets business users with
visual OWL/SHACL creation and model-driven UIs. Schemex~\cite{sadeh2024schemex}
supports iterative schema discovery through AI-assisted abstraction and
contrastive refinement, enabling non-experts to discover structural
patterns from examples with human-in-the-loop validation.
Despite these advances, all existing visual tools are designed either
for ontology specialists or for general non-technical users-none
explicitly address the communication dynamics of cross-functional AI
product teams.

Prot\'eg\'e~\cite{musen2015protege} continues to be the dominant expert
ontology editor but is primarily engineered for ontology specialists
rather than business or domain users, which is repeatedly noted in the
literature.

\subsection{Gaps Addressed by GOI}

The literature reveals six concrete gaps that GOI addresses:

\begin{enumerate}[leftmargin=*]
  \item \textbf{Cross-document canonicalization:} Existing systems
    struggle with unresolved semantically duplicate entities and
    relations across documents~\cite{lairgi2024itext2kg,wang2024autoclusre,zhang2024edc}.
    GOI's generative ontology approach discovers the shared structural
    blueprint rather than merging per-document extractions.
  \item \textbf{Universal typed graph representation:} No standardized
    node and edge taxonomy exists across
    systems~\cite{lo2024endtoend,bai2024autoschemakg}. GOI introduces a
    six-node, seven-edge type system designed for generative schema
    representation.
  \item \textbf{Evaluation for generation tasks:} Token-level metrics
    miss structural correctness~\cite{lo2024endtoend}. GOI's Node
    Coverage Score directly measures whether induced ontologies capture
    the dimensions needed for generation.
  \item \textbf{Interactive visualization for non-experts:} Research
    prototypes include visualization modules but lack production-quality
    interfaces~\cite{lairgi2024itext2kg,abolhasani2025ontokgen,hemid2024kgraphx}.
    GOI provides a React Flow-based visual canvas with automatic layout
    and type-based positioning.
  \item \textbf{Pipeline-ready export:} One-click export in formats
    ready for downstream systems is not yet a first-class
    feature~\cite{lairgi2024itext2kg,tudorache2013webprotege}. GOI provides
    YAML/JSON export designed for RAG and agent integration.
  \item \textbf{Cross-functional team communication:} Existing ontology
    tools are designed either for ontology
    specialists~\cite{musen2015protege} or for researchers, with no
    attention to the communication needs of mixed-expertise engineering
    teams~\cite{hemid2024kgraphx,tudorache2013webprotege,sadeh2024schemex}.
    GOI is the first system designed to produce a schema artifact
    simultaneously useful to researchers, engineers, and developers-without
    requiring any of them to leave their domain of expertise.
\end{enumerate}

\section{The GOI Framework}
\label{sec:framework}

\subsection{Problem Formulation}

Let $\mathcal{D} = \{d_1, d_2, \ldots, d_n\}$ be a corpus of $n$
documents belonging to the same document class $C$ (e.g., all documents
are research papers, or all are job postings, or all are product
specifications). The \textbf{Generative Ontology Induction} problem is
to discover a typed graph $\mathcal{G} = (V, E, \tau_V, \tau_E)$ where:
\begin{itemize}
  \item $V$ is a set of nodes representing schema elements,
  \item $E \subseteq V \times V$ is a set of directed edges,
  \item $\tau_V : V \rightarrow \mathcal{T}_V$ assigns each node a type
    from a fixed node type vocabulary $\mathcal{T}_V$,
  \item $\tau_E : E \rightarrow \mathcal{T}_E$ assigns each edge a type
    from a fixed edge type vocabulary $\mathcal{T}_E$,
\end{itemize}
such that $\mathcal{G}$ represents the \emph{generative blueprint} of
class $C$-the structural schema that governs how documents of type $C$
are composed and that enables generation of new instances of $C$ from a
fresh prompt.

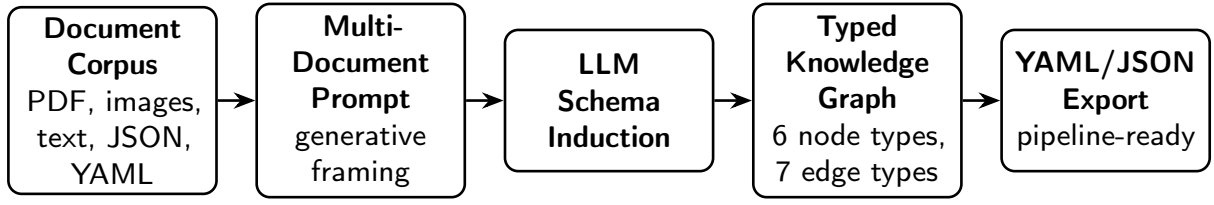
\begin{figure}[t]
\centering
\resizebox{\textwidth}{!}{%
\begin{tikzpicture}[
  node distance=4mm,
  every node/.style={
    rectangle, rounded corners, draw, thick, align=center,
    minimum height=15mm, text width=20mm, font=\sffamily\small
  },
  arr/.style={-Stealth, thick}
]
\node (corpus) {\textbf{Document}\\\textbf{Corpus}\\\footnotesize PDF, images,\\\footnotesize text, JSON, YAML};
\node[right=of corpus] (prompt) {\textbf{Multi-Document}\\\textbf{Prompt}\\\footnotesize generative\\\footnotesize framing};
\node[right=of prompt] (llm) {\textbf{LLM}\\\textbf{Schema}\\\textbf{Induction}};
\node[right=of llm] (graph) {\textbf{Typed}\\\textbf{Knowledge Graph}\\\footnotesize 6 node types,\\\footnotesize 7 edge types};
\node[right=of graph] (export) {\textbf{YAML/JSON}\\\textbf{Export}\\\footnotesize pipeline-ready};
\draw[arr] (corpus) -- (prompt);
\draw[arr] (prompt) -- (llm);
\draw[arr] (llm) -- (graph);
\draw[arr] (graph) -- (export);
\end{tikzpicture}%
}
\caption{GOI's end-to-end ontology induction pipeline. Multi-format
documents are sampled into a single multi-document prompt that
elicits a typed graph; the resulting structure exports directly to
YAML/JSON for downstream LLM pipelines.}
\label{fig:pipeline}
\end{figure}

This formulation distinguishes GOI from entity extraction (which
produces per-document instance graphs) and from ontology population
(which adds instances to a predefined schema). GOI operates at the
\emph{class level}: it discovers the schema, not instances.
Figure~\ref{fig:pipeline} shows the end-to-end pipeline.

\subsection{Node and Edge Type System}

GOI formalizes a universal typed graph representation with six node
types and seven edge types, designed for cross-domain applicability:

\paragraph{Node Types ($|\mathcal{T}_V| = 6$):}
\begin{itemize}
  \item \texttt{class} - A major conceptual category in the document
    type (e.g., ``Author'', ``Section'', ``Product'').
  \item \texttt{property} - An attribute of a class (e.g., ``title'',
    ``publication\_date'', ``price'').
  \item \texttt{value} - A specific allowed value or enumeration
    member for a property.
  \item \texttt{dimension} - A structural axis or organizing principle
    of the document type (e.g., ``Methodology'', ``Results'',
    ``Responsibilities'').
  \item \texttt{relation} - A named relationship between classes.
  \item \texttt{constraint} - A rule or restriction on values,
    cardinalities, or co-occurrence.
\end{itemize}

{\sloppy\paragraph{Edge Types ($|\mathcal{T}_E| = 7$):}
\texttt{is\_a}, \texttt{has\_property}, \texttt{has\_value},
\texttt{relates\_to}, \texttt{part\_of}, \texttt{constrains},
\texttt{instance\_of}.\par}

This type system is deliberately domain-neutral. The same six node types
and seven edge types apply equally to a corpus of scientific papers, job
postings, legal contracts, or product specifications, without any
domain-specific modification.

\subsection{Multi-Document Prompt Architecture}

The core of GOI's induction process is a carefully designed
multi-document prompt that instructs the LLM to reverse-engineer the
generative schema rather than describe individual documents. The system
prompt reads:

\begin{quote}
\textit{``Extract the generative ontology: the set of entities,
dimensions, properties, and relationships that define this kind of
document as a class, so the ontology can be used as a structured
knowledge framework to generate new similar documents from a fresh
prompt.''}
\end{quote}

Multiple input documents are injected as labeled examples
(\texttt{<example\_1>}, \texttt{<example\_2>}, etc.) within a single
prompt context. This multi-example grounding serves two functions: (1)
it forces the LLM to abstract across instances rather than describe any
single document, and (2) it reduces hallucination by anchoring schema
elements in observed patterns across the corpus.

The system accepts documents in multiple formats: PDF (parsed to text),
images (via vision model), plain text, JSON, and YAML. Each document is
preprocessed and labeled before injection into the prompt context. For
very large corpora, documents are sampled to fit within token budget
constraints.

\subsection{Automatic Layout Algorithm}

After schema induction, GOI applies an automatic layout algorithm
(\texttt{layoutNodes()}) that positions nodes by type in horizontal
rows, ensuring immediate visual comprehension without manual
arrangement. Node types are assigned to rows in a fixed vertical order:
\texttt{dimension} nodes at the top (structural axes), followed by
\texttt{class}, \texttt{relation}, \texttt{property}, \texttt{value},
and \texttt{constraint} nodes. Within each row, nodes are distributed
horizontally with uniform spacing. This layout convention encodes
semantic hierarchy visually: the most structurally significant elements
(dimensions) appear at the top, while terminal elements (values,
constraints) appear at the bottom.

\subsection{Import and Export}

GOI supports import of existing ontologies in OWL, RDF, YAML, JSON,
Markdown, and plain text formats, enabling integration with existing
knowledge engineering workflows. The system converts imported
representations into the six-node, seven-edge typed graph format for
visualization and editing.

Export is available in YAML and JSON. The exported document contains
three top-level views. The first is ontology metadata (name,
description, domain). The second is a raw \emph{graph view} that
preserves the typed-node and typed-edge structure for round-trip
editing. The third is a denormalized \emph{pipeline view} in which
each class is paired with its resolved properties, relations, parent,
and examples; each dimension with its enumerated values; and each
constraint with the classes it applies to. A \texttt{promptReady}
field renders the ontology as a markdown system-prompt block, allowing
direct injection into RAG and agent pipelines without any graph
traversal on the consumer side. This dual-view design separates the
round-trip representation from the consumer-facing representation:
graph editors and re-import paths read the graph view, while LLM
prompts and data-validation pipelines read the pipeline view.

\subsection{Interactive Visualization and Editing}

\begin{figure}[t]
  \centering
  \includegraphics[width=\textwidth]{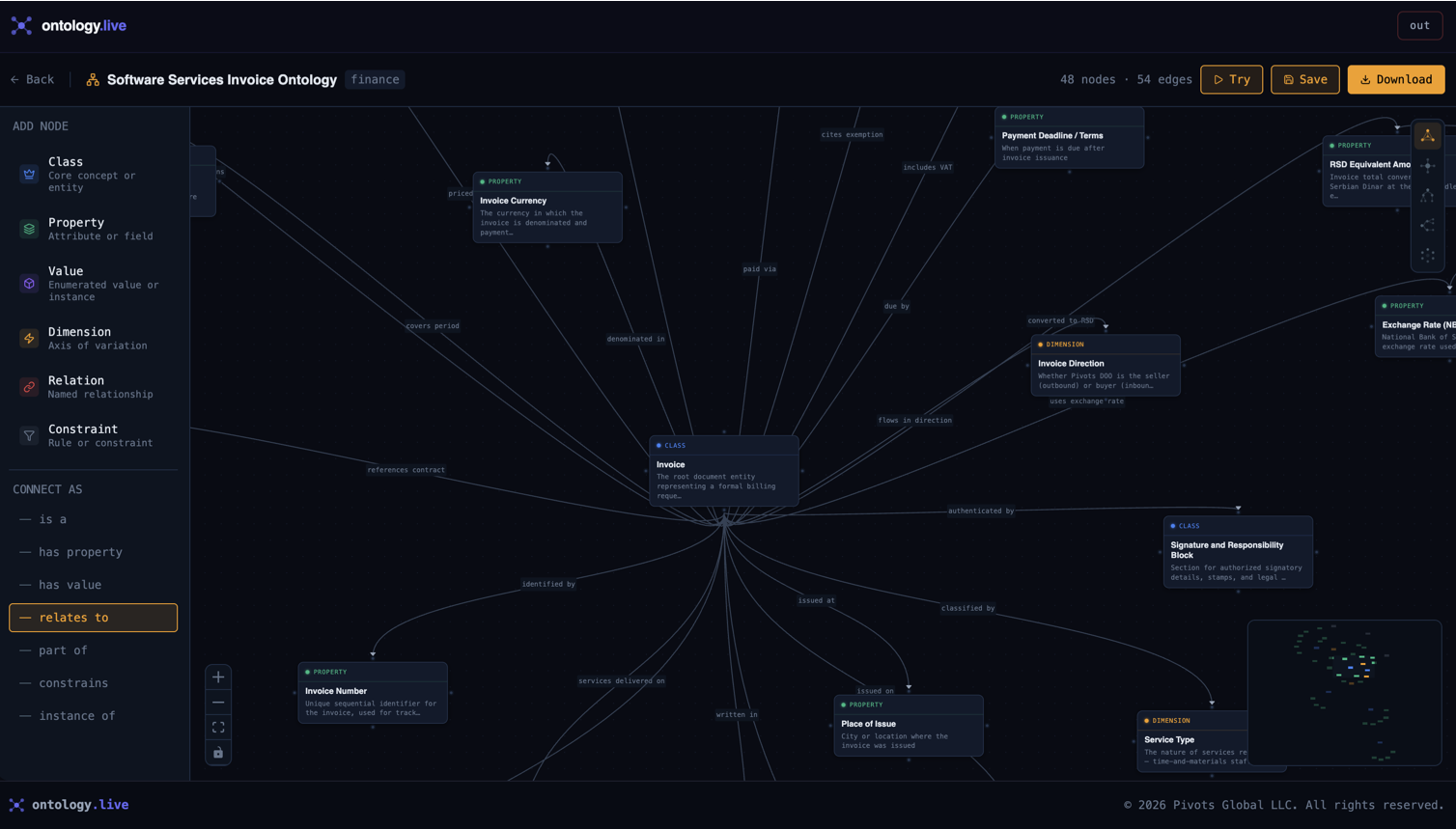}
  \caption{The \emph{Software Services Invoice Ontology} rendered in the
    \texttt{ontology.live} editor (48 nodes, 54 edges). The left
    palette enumerates GOI's typed vocabulary - class, property,
    value, dimension, relation, constraint - and the typed edge set
    used to connect them (\texttt{is\_a}, \texttt{has\_property},
    \texttt{has\_value}, \texttt{relates\_to}, \texttt{part\_of},
    \texttt{constrains}, \texttt{instance\_of}). The central
    \texttt{Invoice} class anchors the graph, with seller, buyer,
    line-item, and bank sub-classes radiating outward; dimension
    nodes (Invoice Direction, Service Type) and constraints
    surface the cross-cutting axes that govern the document type.}
  \label{fig:editor}
\end{figure}

The GOI interface is built on React Flow, providing a drag-and-drop
canvas for visual exploration and editing of induced schemas
(Figure~\ref{fig:editor}). Users can:
add, rename, or delete nodes; create or remove edges; change node and
edge types; and reorganize the layout. The canvas supports zoom,
pan, and node search. All edits are persisted in a SQLite database via
\texttt{better-sqlite3}.

\subsection{System Architecture}

The reference implementation is a TypeScript web application backed
by a relational store, with Anthropic's Claude as the LLM backend.

\section{The Node Coverage Score}
\label{sec:metric}

\subsection{Motivation}

Existing evaluation metrics for ontology quality focus on token-level
accuracy (precision, recall, F1 against reference ontologies) or
structural distance measures~\cite{lo2024endtoend}. These metrics assess
whether the induced ontology matches a ground-truth schema but do not
assess whether the ontology is \emph{useful} for its primary downstream
task: generating new instances of the document class.

A generative ontology is useful if, when used as a generation
specification, the output it produces reflects the structural dimensions
it defines. An ontology that defines a ``Methodology'' dimension but
produces outputs with no methodology section has failed its generative
purpose, regardless of how well it matches a reference schema.

\subsection{Formal Definition}

Let $\mathcal{G} = (V, E, \tau_V, \tau_E)$ be an induced ontology and
let $o$ be an output generated from $\mathcal{G}$ as a generation
specification. Let $S \subseteq V$ be the set of \emph{structural
nodes}: nodes whose type is \texttt{class}, \texttt{property}, or
\texttt{dimension}, restricted to those representing visible
section-level slots (nodes flagged \texttt{generate=context} -
generation-control levers such as employer-brand voice or document
tone - are excluded). Value, relation, and constraint nodes are also
excluded: values are leaves consumed by their parent dimensions,
relations are realised as edge annotations, and constraints describe
rules that hold over other nodes rather than discrete content slots.

The \textbf{Node Coverage Score} is defined as:
\[
  \text{NCS}(\mathcal{G}, o) = \frac{|\{s \in S : s \text{ is mentioned in } o\}|}{|S|}.
\]

{\sloppy
In practice, the metric is reported alongside the diagnostic tuple
$(|S|,\ |\{s \in S : s\text{ is reflected in }o\}|,\ \text{NCS})$, plus
the per-node verdict for error analysis. The original
dimension-only formulation reported in earlier drafts is recovered as
the special case of restricting $S$ to dimension nodes.\par}

\subsection{Computation}

The coverage score is computed by serializing the ontology into a
structured prompt, firing a user prompt through the ontology as a
generation specification, and then scanning the output for mentions of
each structural node. A node $s$ is considered ``mentioned'' if any of
six detectors fires: full-label substring match, single-token label
match, all-token label match, distinctive-token match (a single label
token of $\geq 5$ characters present in the output, used for
compositional axis labels), example-value match, or per-case
synonym match. Detectors are evaluated case-insensitively; the
synonym table is published with the supplementary materials so the
score is reproducible.

The metric supports dataset injection for data-driven generation,
allowing evaluation under realistic conditions where the generation
prompt includes actual data from the target domain.

\subsection{Limitations of the Metric}

The Node Coverage Score has several limitations. It measures
structural completeness (presence of class, property, and dimension
slots) and does not assess constraint satisfaction or content
correctness within each slot. Synonym tables and distinctive-token
matching introduce ambiguity in borderline cases. The metric does not
measure the quality or accuracy of the content within each covered
node, only its presence. Despite these limitations, the Node Coverage
Score provides a novel evaluation axis that complements token-level
metrics and directly assesses ontology utility for generation tasks.

\section{Experiments and Evaluation}
\label{sec:experiments}

We evaluate GOI through a controlled generative validation: given an
ontology as the only context, we measure whether an LLM can instantiate
a complete, schema-valid document from scratch. The ontology itself
serves as the gold standard - a property that holds independently of
any labeled corpus.

\subsection{Corpora}
\label{subsec:corpora}

The Software Services Invoice ontology used below was induced from a
corpus of 20 real B2B invoices internal to the author's
organization; this corpus is withheld from supplementary materials due
to seller, client, and contractor confidentiality, but a
structurally-identical anonymized version of the resulting ontology is
included. The Job Description Ontology was induced from a
publicly accessible corpus of senior engineering job postings; three
anonymized postings are reproduced in the supplementary bundle.
The Pain-Management Clinical Visit Record Ontology was induced from a
confidential corpus of pain management and rehabilitation follow-up
visit records held by a clinical partner of the author's
organization. The corpus contains protected health information (PHI)
and is withheld in full; the published ontology JSON has every
example value, provider name, practice name, address, patient
identifier, and signature date replaced with synthetic
clinically-plausible equivalents, so the structural schema is
reproduced faithfully without any real PHI.
The Professional Services Contract \& Statement of Work Ontology was
induced from a confidential corpus of 20 consulting agreements,
contractor contracts, and statements of work internal to the author's
organization. The corpus carries party-identifying information
(client and provider legal names, registered addresses, tax IDs,
contact details, agreement identifiers, and authorized signatories)
and is withheld in full; the published ontology JSON has every such
example value replaced with synthetic legally-plausible equivalents
(synthetic Serbian and Western corporate names, plausible-but-fictional
addresses, synthetic registration / tax-ID numbers of similar
format, synthetic contact information under the new corporate domain,
synthetic agreement identifiers preserving the original
project-code-plus-date pattern, and synthetic signatory names with
matching role/title structure), so the structural schema is reproduced
faithfully without any real party-identifying data.

\subsection{Ontology-as-Prompt: Generative Validation}
\label{sec:ontology-as-prompt}

A core question for any induced ontology is whether it is dense enough
to \emph{originate} a document-whether, supplied as the only context,
it lets an LLM instantiate a complete, schema-valid artifact from
scratch. The gold standard (the ontology itself) coincides with the
input, so this evaluation is intrinsically reproducible: the ontology
defines its own coverage contract.

\medskip\noindent\textbf{Protocol.} For each ontology, \textbf{System~A} is the GOI
generation prompt: the full GOI ontology JSON with an instruction to
emit a JSON map keyed by ontology node id, instantiating every class,
populating every property, reflecting each dimension, and satisfying
every constraint. \textbf{System~B} is a generic 3-field template
(\texttt{title}, \texttt{content}, \texttt{metadata}). Both systems
run on the same LLM (Anthropic Claude Sonnet, fixed model identifier
\texttt{claude-sonnet-4-6}) with identical decoding settings; no
corpus is consumed-generation is from scratch in both cases. We
evaluate four ontologies of contrasting familiarity: a Software
Services Invoice ontology (a document type the LLM has seen at scale
during pretraining), a custom Job Description Ontology (with
non-standard dimensions and explicit context-vs-content separation),
a Pain-Management Clinical Visit Record Ontology (workers'
compensation and musculoskeletal pain documentation, with a highly
specialised structural backbone covering claim type, body region,
exam-by-region modules, ICD-10 assessment, and treatment
justification), and a Professional Services Contract \& Statement
of Work Ontology (consulting agreements and SOWs, with structural
slots for parties and signatories, hourly-vs-fixed fee structures
with weekly hours caps, milestone tables, IP assignment direction,
non-solicitation, exclusivity, indemnification, and SOW-specific
sections such as context/background, assumptions and risks,
acceptance criteria, and communication plan). Full verbatim prompts
are included in the supplementary bundle.

\medskip\noindent\textbf{Results.} Table~\ref{tab:generative_validation} summarises
the four cases.

\begin{table}[!htbp]
\centering
\caption{Generative validation: structural-node coverage (classes,
properties, and dimensions; non-context only) of LLM-generated
documents when prompted with the full ontology (System~A) versus a
generic three-field template (System~B). Familiar document types
yield small coverage gaps because the model's pretraining prior fills
in expected fields; custom and specialised ontologies with
non-standard dimensions widen the gap dramatically. All four cases
scored under the broadened metric described in
\S\ref{sec:metric}.}
\label{tab:generative_validation}
\begin{tabularx}{\textwidth}{@{}>{\centering\arraybackslash}X >{\raggedright\arraybackslash}X >{\centering\arraybackslash}X >{\centering\arraybackslash}X >{\centering\arraybackslash}X >{\centering\arraybackslash}X@{}}
\toprule
\textbf{Case} & \textbf{Ontology} & \textbf{Nodes} & \textbf{GOI (A)} & \textbf{GST (B)} & \textbf{Gap} \\
\midrule
1 & Software Services Invoice & 45 & \textbf{97.8\%} & 97.8\% & 0.0~pp \\
2 & Job Description Ontology & 23\textsuperscript{*} & \textbf{100.0\%} & 52.2\% & 47.8~pp \\
3 & Pain-Mgmt Clinical Visit & 45 & \textbf{95.6\%} & 62.2\% & 33.4~pp \\
4 & Prof.\ Services Contract \& SOW & 46 & \textbf{100.0\%} & 78.3\% & 21.7~pp \\
\bottomrule
\end{tabularx}
\\[2pt]
{\footnotesize \textsuperscript{*}Applicable non-context structural
nodes per the ontology's own coverage contract (47 structural nodes
total; 24 flagged \texttt{generate=context}; 23 applicable to a
single JD instance).}
\end{table}

\medskip\noindent\textbf{Case~1 (familiar).} On the Software Services Invoice ontology
(45 structural nodes), System~A instantiated 44/45 nodes (97.8\%),
including all six classes, all four dimensions, and all three
constraints with verified arithmetic (line totals sum to invoice
total; sales-tax exemption note for out-of-state buyer;
foreign-currency conversion trivially satisfied for USD-only); the
single uninstantiated node was an exchange-rate field that is
legitimately N/A for the chosen US-domestic instance. System~B
instantiated 44/45 nodes (97.8\%), with all four dimensions appearing
as incidental \texttt{metadata} keys driven by the model's strong
invoice prior. The gap closes to zero because the LLM has seen enough
invoices during pretraining to recover essentially the entire
structural backbone incidentally even from a generic template.

\medskip\noindent{\sloppy\textbf{Case~2 (custom).} On the Job Description Ontology
(63 nodes total; 47 structural; 24 flagged \texttt{generate=context};
23 applicable structural nodes per the ontology's own coverage
contract), System~A instantiated 23/23 applicable non-context nodes
(100\%). System~B instantiated only 12/23 (52.2\%): it omitted the
\texttt{cat\_org\_context} class entirely (no About-the-Company
section was emitted) and missed many custom
properties-\texttt{company\_stage}, \texttt{employer\_brand},
\texttt{career\_trajectory}, and
\texttt{certification\_profile} among them. The gap widens to
47.8~pp because the LLM has no prior for taxonomy-specific dimensions
like employer-brand voice, candidate-market targeting,
career-trajectory pattern, or certification-profile expectations;
without an explicit schema, these axes simply do not surface.\par}

\medskip\noindent\textbf{Case~3 (specialised, anonymized).} On the Pain-Management
Clinical Visit Record Ontology (45 structural nodes), System~A
instantiated 43/45 nodes (95.6\%); the two uninstantiated nodes
(\texttt{Signature Block} and \texttt{Radiologist}) were lost to
output-token truncation at the end of an unusually long generation,
not to a structural omission. System~B emitted a primary-care
follow-up note for hypertension and type 2 diabetes-it correctly
inferred a clinical visit document but missed the workers'
compensation and pain-management specialisation entirely-and
covered only 28/45 structural nodes (62.2\%). System~B omitted the
entire claim-information sub-graph (\texttt{Claim Type},
\texttt{Causal Relationship}, \texttt{Work Status},
\texttt{Temporary Impairment Percentage}), the regional physical-exam
modules (\texttt{Range of Motion}, \texttt{Special Orthopedic Tests},
\texttt{Sensation Testing}, \texttt{Muscle Stretch Reflexes}), and
the recommendations apparatus (\texttt{Treatment Justification},
\texttt{Functional Goals}, \texttt{ADL Impact},
\texttt{Radiology Report}). The 33.4~pp gap reflects how much
structural specialisation the ontology contributes in a
healthcare-documentation context where the LLM's prior reverts to
the generic primary-care template.

\medskip\noindent\textbf{Case~4 (legal, anonymized).} On the Professional Services
Contract \& Statement of Work Ontology (46 structural nodes),
System~A instantiated 46/46 nodes (100\%), including all subclasses
and relations across the parties block, the milestone table, the IP,
confidentiality, non-solicitation, warranties, termination,
indemnification, and assignability clauses, the SOW-specific sections
(context/background, assumptions and risks, acceptance of
deliverables, communication plan), the hourly-rate fee structure
with weekly hours cap, and the dispute-resolution mechanism with
explicit JAMS arbitration. System~B emitted a generic fixed-fee UX
design contract and covered 36/46 structural nodes (78.3\%); the gap
is narrower than Cases~2 and~3 because professional-services
contracts share substantial universal vocabulary with the LLM's
training prior (governing law, termination, IP assignment,
confidentiality, entire-agreement). The 10 missed nodes are
precisely the SOW-specific structural backbone the LLM's prior does
not surface unprompted: the \texttt{Auto-Renewal Clause},
\texttt{Hourly Rate} (the baseline reverted to a fixed-fee model),
\texttt{Non-Solicitation Clause}, \texttt{Exclusivity Clause},
\texttt{Severability Clause}, \texttt{SOW Number / Reference},
\texttt{Assumptions, Risks and Constraints}, \texttt{Communication
Plan}, \texttt{Weekly Hours Cap}, and \texttt{Indemnification
Clause}. The 21.7~pp gap quantifies the structural specialisation
GOI contributes once a contract goes beyond a boilerplate
fixed-fee engagement and into a milestone-driven SOW with
hours-capped hourly billing and detailed clause-level provisions.

The by-construction vs incidental framing now lands harder: GOI
provides a structural contract that holds \emph{regardless of how
familiar the document type is to the model}, whereas the generic
template's coverage is contingent on the training-data prior and
collapses on novel or specialised structure.

\medskip\noindent\textbf{Implication.} For industrial deployment, the headline
metric is not raw coverage on familiar document types but
\emph{worst-case} coverage on novel or specialised schemas, where
downstream pipelines cannot rely on the LLM's prior to fill the gaps.
GOI's contribution is precisely this structurally-mandated coverage: every
ontology node is either instantiated or explicitly marked
\texttt{null}, by construction. The generative results in
Table~\ref{tab:generative_validation} establish a lower bound on GOI's
structural reliability for the schema-to-document direction.

\section{Discussion}
\label{sec:discussion}

\subsection{Contributions and Implications}

The \textbf{generative-ontology framing} sidesteps cross-document
canonicalization by never producing per-document instance graphs.
The \textbf{six-node, seven-edge type system} supplies a
standardized cross-domain taxonomy where prior work uses
domain-specific schemas or untyped triples. The \textbf{Node
Coverage Score} adds an evaluation axis aligned with structural
completeness rather than token-level accuracy. As a side-effect of
producing a typed graph artifact, GOI's induced ontologies serve as a
shared structural reference across mixed-expertise teams without
privileging any single discipline's vocabulary.

\subsection{Limitations}

GOI has several important limitations. \textbf{LLM consistency:}
non-determinism yields schemas that vary in naming and granularity
across runs; multi-example grounding reduces but does not eliminate
this variance. \textbf{Token constraints:} large corpora must be
sampled to fit context windows, risking loss of patterns present
only in unsampled documents. \textbf{Hallucination:} LLMs may
introduce schema elements absent from the corpus, motivating
human-in-the-loop validation for high-stakes use. \textbf{No formal
OWL axioms:} GOI produces typed JSON rather than OWL/RDF and does
not infer description-logic constraints. \textbf{Limited
relationship discovery:} cardinalities, inverse properties, and
transitive closure are not automatically derived. \textbf{Metric
scope:} Node Coverage measures structural completeness over class,
property, and dimension nodes but does not assess constraint
satisfaction, property-value correctness, or content quality within
each covered slot.

\section{Conclusion}
\label{sec:conclusion}

\textbf{Generative Ontology Induction (GOI)} reverse-engineers the
structural schema of a document class into a six-node, seven-edge
typed graph with YAML/JSON export. Across four contrasting
ontologies in four distinct domains-B2B billing, talent acquisition,
healthcare clinical documentation, and professional services
contracting-our controlled generative validation shows GOI-prompted
generation covers 95-100\% of the structural backbone in every case.
A generic three-field template matches GOI on the familiar invoice
schema (97.8\%) but collapses to 52.2\% on the custom Job Description
Ontology, 62.2\% on the Pain-Management Clinical Visit Record
Ontology, and 78.3\% on the Professional Services Contract \&
Statement of Work Ontology-a structural guarantee generic prompting
cannot offer once the document type leaves the LLM's pretraining
prior.

\section*{Acknowledgments and Data Availability}
This research received no external funding. Source code, a live
demonstration of the reference implementation, video walkthroughs,
and an archival snapshot are available at the following links:
\begin{itemize}
  \setlength{\itemsep}{0pt}
  \item Live demo: \url{https://ontology.live}
  \item Source code: \url{https://github.com/cjsergienko/ontology-open}
  \item Video walkthroughs: \url{https://www.youtube.com/@GenerativeOntologyInduction}
  \item Archival snapshot (Zenodo): \url{https://doi.org/10.5281/zenodo.19893755}
\end{itemize}

\paragraph{Supplementary Materials.}\sloppy A reproducibility bundle at
\texttt{research/supplementary/} contains the Job Description
Ontology JSON (63 nodes), an anonymized structural version of
the Software Services Invoice ontology JSON (48 nodes; example
values replaced with synthetic placeholders), an anonymized
Pain-Management Clinical Visit Record Ontology JSON (45 nodes;
provider, practice, address, patient, and date values replaced with
synthetic clinically-plausible equivalents), an anonymized
Professional Services Contract \& Statement of Work Ontology JSON
(46 nodes; party legal names, addresses, registration / tax-ID
numbers, contact information, agreement identifiers, and signatory
names replaced with synthetic legally-plausible equivalents), the
eight LLM-generated documents used in
\S\ref{sec:ontology-as-prompt}, the two prompt templates (System~A
and~B), and the broadened-NCS
scoring protocol with per-case detection log. Three publicly-listed
job postings used to induce the JD ontology are at
\texttt{research/supplementary/job\_descriptions/}.\par

\bibliographystyle{splncs04}

\end{document}